\documentclass[10pt,twocolumn,letterpaper]{article}

\usepackage{iccv}
\usepackage{times}
\usepackage{epsfig}
\usepackage{graphicx}
\usepackage{amsmath}
\usepackage{amssymb}
\usepackage{orcidlink}
\usepackage{float}

\newcommand{\x}{\mathbf{x}}
\newcommand{\s}{\mathbf{s}}
\newcommand{\z}{\mathbf{z}}
\newcommand{\pxs}{p_\text{NM}(\x|\s)}
\newcommand{\psx}{p(\s|\x)}
\newcommand{\pz}{p_\theta(\z)}
\newcommand{\pxz}{p_\theta(\x|\z)}
\newcommand{\px}{p_\theta(\x)}
\newcommand{\qzx}{q_\phi(\z|\x)}
\newcommand{\pzx}{p_\theta(\z|\x)}

\iccvfinalcopy % *** Uncomment this line for the final submission

\def\iccvPaperID{1} % *** Enter the ICCV Paper ID here

\ificcvfinal\fi

\newcommand\figexpl{
\begin{figure*}[h!]
    \centering
    \includegraphics[width=1\linewidth]{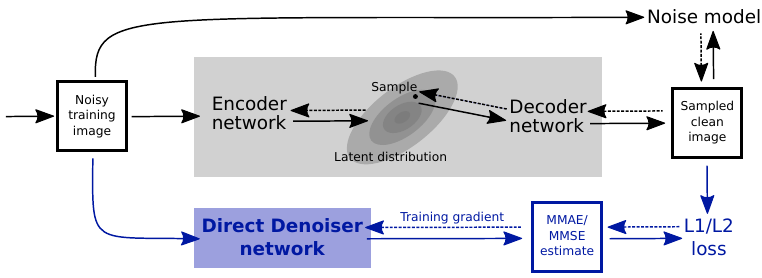}
    \caption{
    \textbf{Training scheme:} 
    We train our novel \emph{Direct Denoiser} (blue) along side a Variational AutoEncoder (VAE)~\cite{prakash2020fully, prakash2022interpretable}.
    The processing of data is shown with solid arrows and the backward propagation of gradients required for training is shown with dashed arrows.
    The VAE encoder takes a noisy image as input and predicts the parameters of a distribution in latent space, a sample is drawn from here and mapped to a possible clean image by the decoder network.
    The reconstruction loss is computed using a pre-trained noise model.
    Our Direct Denoiser is trained using noisy images as input and the clean image samples (predicted by the VAE) as target.
    Since individual samples differ for the same input, there is no unique correct solution for this task.
    As a consequence, by using an $L_2$ loss, the Direct Denoiser will learn to predict the expected value, \ie, the MMSE solution.
    Using an $L_1$ loss leads to predicting the pixel-wise median.
    We block gradients from passing through the sampled clean image to prevent the VAE changing its outputs.
    }
    \label{fig:explainer}
\end{figure*}
}

\newcommand\tabpsnrs{
\begin{table*}[]
\centering
\caption{
\textbf{Average PSNR of consensus solutions from HDN~\cite{prakash2022interpretable} and direct solutions from our novel \emph{Direct Denoiser}.}
HDN's consensus solutions were obtained by taking samples of varying sizes from its denoising distribution and calculating both their per-pixel median and their per-pixel mean.
The Direct Denoiser's solutions were obtained from a single pass of a network trained under an $L_1$ loss and a single pass of a network trained under an $L_2$ loss.
PSNRs are presented as the median/mean consensus for HDN and as the solution from the $L_1$/$L_2$ network for the Direct Denoiser.
Best results are printed in \textbf{bold}.
}
\begin{tabular}{l|cccc|c}
                                      & \multicolumn{4}{c|}{\textbf{Number of samples (HDN)}}               & \multicolumn{1}{l}{} \\
\multicolumn{1}{c|}{\textbf{Dataset}} & \textbf{1}    & \textbf{10}   & \textbf{100}  & \textbf{1000} & \textbf{Direct}               \\ \hline
Convallaria                           & 33.69 / 33.69 & 36.59 / 36.76 & 37.17 / 37.23 & 37.19 / 37.27 & \textbf{37.50} / 37.45        \\
Confocal Mice                         & 35.43 / 35.43 & 37.30 / 37.42 & 37.58 / 37.68 & 37.62 / 37.69 & \textbf{37.77} / 37.75        \\
2 Photon Mice                         & 31.21 / 31.21 & 32.63 / 32.68 & 32.86 / 32.87 & 32.89 / 32.89 & \textbf{33.55} / 33.54        \\
Mouse Actin                           & 31.62 / 31.62 & 33.52 / 33.66 & 33.87 / 33.92 & 33.91 / 33.95 & 34.22 / \textbf{34.28}        \\
Mouse Nuclei                          & 33.48 / 33.48 & 36.24 / 36.44 & 36.79 / 36.89 & 36.81 / 36.90 & 36.87 / \textbf{36.93 }       \\
Struct. Convallaria                   & 29.02 / 29.02 & 30.88 / 31.00 & 31.22 / 31.27 & 31.27 / 31.29 & 31.58 / \textbf{31.64}       
\end{tabular}
\label{tab:PSNRs}
\vspace{-2mm}
\end{table*}
}

\newcommand\figresults{
\begin{figure*}[h!]
    \centering
    \includegraphics[width=1\linewidth]{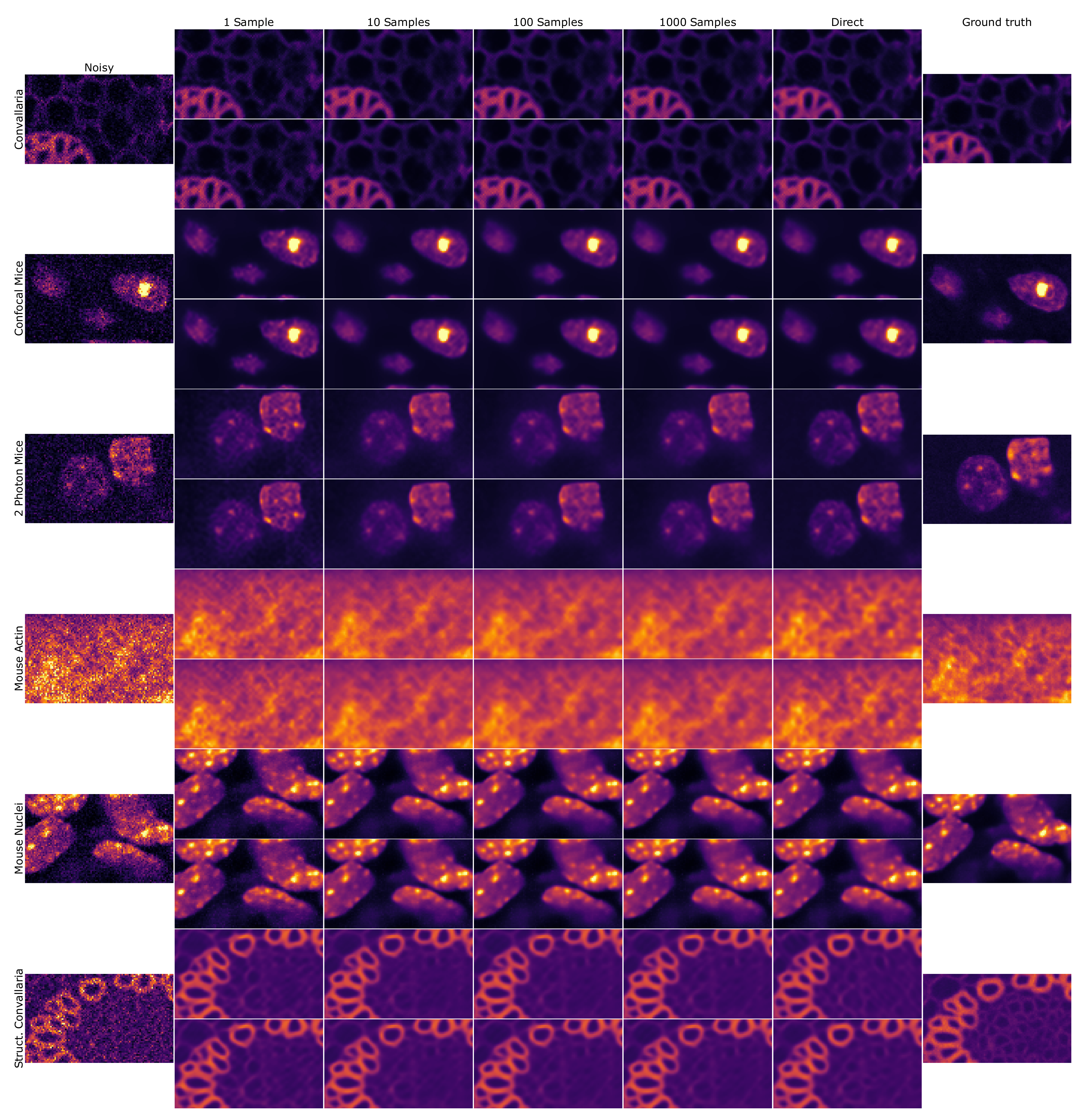}
    \caption{
    \textbf{Visual results:} 
    Cropped images from each dataset showing consensus solutions of varying sample sizes from HDN's denoising distribution with direct solutions from our Direct Denoiser.
    For each dataset, the top row shows the median of HDN samples and a solution from our $L_1$ trained Direct Denoiser, while the bottom row shows the mean of HDN samples and a solution from our $L_2$ trained Direct Denoiser.
    }
    \label{fig:results}
\end{figure*}
}

\newcommand\figinferencetime{
\begin{figure}[h!]
    \hspace{-4mm}
    \includegraphics[width=1\linewidth]{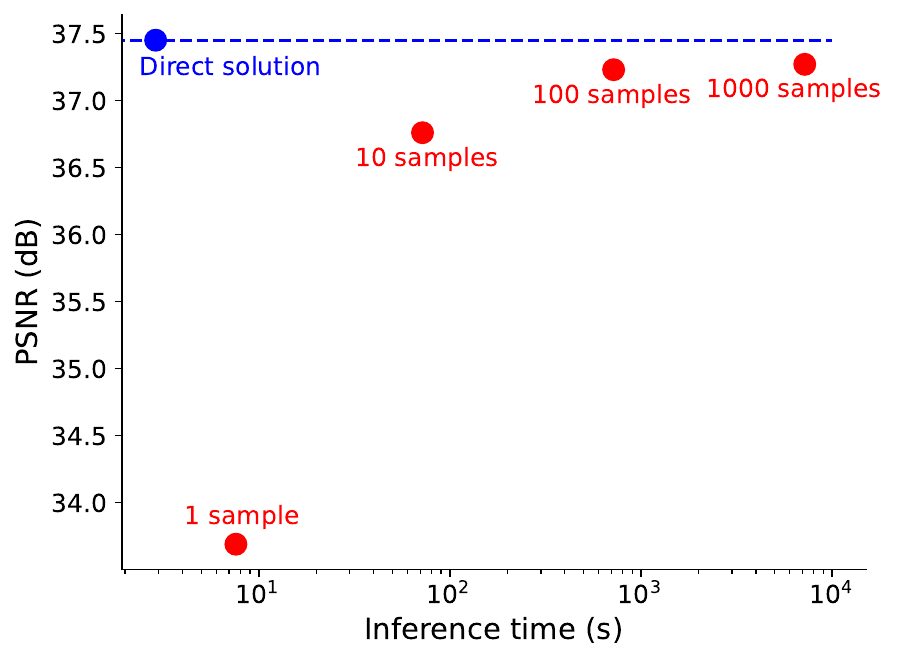}
    \caption{
    \textbf{Our Direct Denoiser outperforms  unsupervised VAE-based denoising (HDN)~\cite{prakash2022interpretable}, while requiring only a fraction of the computational cost:}
    In red, the time to draw 1, 10, 100 and 1000 samples from HDN's learned denoising distribution plotted against the PSNR (higher is better) of the per-pixel mean of these samples.
    Additionally, in blue, the time to take a single solution from our Direct Denoiser is plotted against its PSNR.
    These results are from denoising the \emph{Convallaria} dataset.
    }
    \label{fig:inferencetime}
\end{figure}
}

\begin{document}

%%%%%%%%% TITLE
\title{Direct Unsupervised Denoising}
\author{Benjamin Salmon
\orcidlink{0000-0002-5919-0158} and Alexander Krull\orcidlink{0000-0002-7778-7169}\\
University of Birmingham\\
Birmingham B15 2TT, UK\\
{\tt\small brs209@student.bham.ac.uk, a.f.f.krull@bham.ac.uk}
% For a paper whose authors are all at the same institution,
% omit the following lines up until the closing ``}''.
% Additional authors and addresses can be added with ``\and'',
% just like the second author.
% To save space, use either the email address or home page, not both
% \and
% Second Author\\
% Institution2\\
% First line of institution2 address\\
% {\tt\small secondauthor@i2.org}
}

\maketitle
% Remove page # from the first page of camera-ready.
\ificcvfinal\thispagestyle{empty}\fi

%%%%%%%%% ABSTRACT
% ---------------------------------------------------
\begin{abstract}
% ---------------------------------------------------
Traditional supervised denoisers are trained using pairs of noisy input and clean target images.
They learn to predict a central tendency of the posterior distribution over possible clean images.
%Traditional supervised denoisers are trained using pairs of noisy input and clean target images.
% Depending on the loss function used, these methods predict the central tendency of the posterior distribution over possible clean images.
When, \eg, trained with the popular quadratic loss function, the network's output will correspond to the minimum mean square error (MMSE) estimate.
Unsupervised denoisers based on Variational AutoEncoders (VAEs) have succeeded in achieving state-of-the-art results while requiring only unpaired noisy data as training input.
In contrast to the traditional supervised approach, unsupervised denoisers do not directly produce a single prediction, such as the MMSE estimate, but allow us to draw samples from the posterior distribution of clean solutions corresponding to the noisy input.
To approximate the MMSE estimate during inference, unsupervised methods have to create and draw a large number of samples -- a computationally expensive process -- rendering the approach inapplicable in many situations.
Here, we present an alternative approach that trains a deterministic network alongside the VAE to directly predict a central tendency.
Our method achieves results that surpass the results achieved by the unsupervised method at a fraction of the computational cost.
\end{abstract}

%%%%%%%%% BODY TEXT
% ---------------------------------------------------
\section{Introduction}
% ---------------------------------------------------
\figinferencetime
The prevalence of noise in biomedical imaging makes denoising a necessary step for many applications~\cite{laine2021imaging}. 
Deep learning has proven itself to be the most powerful tool for this task, as is evidenced by a growing body of research~\cite{su2022survey}. 
Although deep learning-based approaches typically require large amounts of training data, recent advances in unsupervised deep learning~\cite{prakash2020fully, prakash2022interpretable, salmon2022towards} have shown that this requirement need not be a barrier to their use.
Unlike with supervised deep learning-based denoisers, which are trained with pairs of corresponding noisy and noise-free images, users of unsupervised methods can train their models with the very data they want to denoise.

The performance of unsupervised deep learning-based denoisers is now approaching and even sometimes matching the performance of their supervised counterparts~\cite{prakash2020fully, prakash2022interpretable, salmon2022towards}, however, these two methods are fundamentally different in the way they do inference.
By training a Variational AutoEncoder (VAE)~\cite{kingma2014auto}, unsupervised methods approximate a posterior distribution over the clean images that could underlie a noisy input image.
This distribution will be referred to as the \emph{denoising distribution}.
% Unlike supervised approaches, unsupervised methods allow us to sample from an approximate posterior distribution over the clean images that could underlie a given noisy input image.
% We will refer to this as the \emph{denoising distribution}.
Random samples from the denoising distribution then constitute the infinite possible solutions to a denoising problem.
Supervised and self-supervised learning methods, on the other hand, offer a single prediction that compromises between all possible solutions.
This is usually a central tendency of the denoising distribution and the specific central tendency that is predicted depends on the loss function used. For example, a supervised method trained with the mean squared error (MSE) loss function will predict the mean, which is also known as the minimum mean squared error (MMSE) estimate.
A model trained with the mean absolute error (MAE) loss function will predict the pixel-wise median, which is known as the minimum mean absolute error (MMAE) estimate.

While the ability of unsupervised methods to produce diverse solutions can in some circumstances be beneficial for downstream processing~\cite{prakash2020fully}, users oftentimes require only a single solution such as the MMSE estimate.
If they are to approximate this from an unsupervised learning-based denoiser, they must process their image many times and average many possible sampled solutions, leading to a significant computational overhead.
For example, the authors of~\cite{prakash2020fully, prakash2022interpretable, salmon2022towards} average 100 or 1000  samples per image to obtain their MMSE estimate.
Such an approach requires substantial computational effort, and is not likely to be economically and ecologically reasonable for labs regularly analyzing terabytes of data.

This paper presents an alternative route to estimating the central tendencies from an unsupervised denoiser; one that requires noisy images to be processed only once.
We do so by training an additional deterministic convolutional neural network (CNN), termed \emph{Direct Denoiser}, that directly predicts MMSE or MMAE solutions and is trained alongside the VAE.
It uses noisy training images as input and the sampled predictions from the VAE as training targets.
Lacking a probabilistic nature, this network will minimize its MSE or MAE loss function by predicting the mean or pixel-wise median of the denoising distribution.
The result is a denoising network with the evaluation times of a supervised approach and the training data requirements of an unsupervised approach.
% We also show that this additional network can be trained alongside the main denoiser and report the effect on training times in section~\ref{sec:experiments}.

In summary, we propose an extension to unsupervised deep learning-based denoisers that dramatically reduces inference time by estimating a central tendency of the learned denoising distribution in a single evaluation step.
Moreover, we show these estimates to be more accurate than those obtained by averaging even up to 1000 samples from the denoising distribution.
Figure~\ref{fig:inferencetime} shows how much shorter inference time is with our proposed approach, and how much higher the quality of results are.

The remainder of the paper is structured as follows.
In Section~\ref{sec:relwork}, we give a brief overview of related work, concentrating on different approaches to denoising.
In Section~\ref{sec:unsuperviseddenoising}, we provide a formal introduction to the unsupervised VAE-based denoising approach, which is the foundation of our method.
In Section~\ref{sec:method}, we describe the training of the Direct Denoiser.
We evaluate our approach in Section~\ref{sec:experiments}, showing that we consistently outperform our baseline at a fraction of the computational cost.
Finally, in Sections~\ref{sec:discussion}~and~\ref{sec:conclusions} we discuss our results and give an outlook on the expected impact of our work and future perspectives.

% ---------------------------------------------------
\section{Related Work}
\label{sec:relwork}
% ---------------------------------------------------

% % ...................................................
% \subsection{Non-learning-based methods}
% % ...................................................

% ...................................................
\subsection{Supervised denoising}
% ...................................................
Traditional supervised deep learning-based methods (\eg \cite{Zhang2017_DnCNN, weigert2018content}) rely on paired training data consisting of corresponding noisy and clean images.
These methods view denoising as a regression problem, and usually train a UNet~\cite{ronneberger2015u} or variants of the architecture to learn a mapping from noisy to clean.
The most commonly used loss function for this purpose is the sum of pixel-wise quadratic errors ($L_2$ or MSE), which directs the network to predict the MMSE estimate for the noisy input.

The approach's requirement for clean training images greatly limits its applicability, particularly for scientific imaging applications, where often no clean data can be obtained. 
In 2018, Lehtinen~\etal~\cite{lehtinen2018noise2noise} had the insight that training of equivalent quality can be achieved by replacing the clean training image with a second noisy image of the same content; a training method termed \emph{Noise2Noise}.
In practice, such image pairs can often be acquired by recording two images in quick succession.
By using the $L_2$ loss and assuming that the imaging noise is zero-centered, the network is expected to minimize the loss to its noisy training target by converging to the same MMSE estimate as in supervised training.

While Noise2Noise and traditional supervised methods are state-of-the-art with respect to the quality of their results, their requirement for paired training data makes them inapplicable in many situations.
In contrast, our method requires only unpaired noisy data, which is available for any denoising task, making it directly applicable in situations where supervised methods are not.

% ...................................................
\subsection{Self-supervised denoising}
% ...................................................
Self-supervised methods have been introduced to enable denoising with unpaired noisy data.
Here we focus on \emph{blind-spot} approaches~(\eg~\cite{krull2019noise2void, batson2019noise2self, moran2020noisier2noise, quan2020self2self}), which mask individual pixels in the input image and use them as training targets.
These methods rely on the assumption that imaging noise is pixel-wise independent given an underlying signal. 
By effectively forcing the network to predict each pixel value from its surroundings, blind-spot approaches can learn to denoise images without the need for paired noisy-clean data.
Like supervised methods, self-supervised denoisers (when used with $L_2$ loss) predict an MMSE estimate for each pixel, albeit based on less information, since the corresponding input pixel cannot be used during prediction.
As a result, the quality of the output can be worse than supervised methods.
The blind-spot approach has been improved to reintroduce the lost pixel information during inference~\cite{Prakash2019ppn2v, laine2019high}, achieving improved quality in some situations.
In \cite{Broaddus2020_ISBI}, Broaddus~\etal extended the method to allow for the removal of structured noise.

Our method also does not require paired data, but we do not follow the self-supervised blind-spot paradigm. 
As a consequence, we do not have to address the loss of pixel information.

% ...................................................
\subsection{Unsupervised VAE-based denoising}
% ...................................................
Unsupervised VAE-based denoising methods~\cite{prakash2020fully} form the backbone of our method. 
Like in self-supervised methods, training requires only noisy images.
However, their training and inference procedures differ greatly from self-supervised approaches.
We discuss this class of methods in detail in Section~\ref{sec:unsuperviseddenoising}.

% ...................................................
\subsection{Knowledge distillation}
% ...................................................
Knowledge distillation~\cite{hinton2015distilling} is the process of training a smaller \emph{student} network using a large \emph{teacher} network or an ensemble~\cite{dietterich2000ensemble} of teachers.
The goal of this approach is to reduce the computational effort required during inference and enable more efficient employment of a powerful model.
Surprisingly, the student model can achieve better results compared to being trained on the data directly.
A survey of the topic can be found in~\cite{gou2021knowledge}.

The approach of training our Direct Denoiser with the output of another network can be seen as knowledge distillation.
However, in our case the Direct Denoiser is not intended as a smaller replacement of the VAE, but as a model with a faster inference procedure.

% HDN. Describe how it works with noise model and everything.
% Noise2Noise
% Maybe PixelHDN.
% Maybe CARE
% Maybe noise2void/self etc.

% ---------------------------------------------------
\section{Background}
\label{sec:unsuperviseddenoising}
% ---------------------------------------------------

% ...................................................
\subsection{The denoising task}
% ...................................................
A noisy observation, $\x$, of a signal, $\s$, can be thought of as sampled from an observation likelihood, or \emph{noise model}, $\pxs$.
A noise model describes the random, unwanted variation that is added to a signal when it is recorded.
The goal of denoising is to estimate the $\s$ that parameterized the noise model from which a known $\x$ was sampled.

% ...................................................
\subsection{Unsupervised denoising}
% ...................................................
It was Prakash~\etal~\cite{prakash2020fully} who proposed doing so via variational inference, using a VAE~\cite{kingma2014auto} to approximate the posterior distribution $\psx$. 
They improved their approach with a more powerful architecture that could also handle mild forms of structured noise in~\cite{prakash2022interpretable}.
Salmon and Krull then presented an alternative approach to tackling structured noise in~\cite{salmon2022towards}, but it unfortunately cannot yet be applied in realistic settings.

To understand how unsupervised denoising works, we must give a brief explanation of the VAE~\cite{kingma2014auto}.
For a full introduction, see~\cite{doersch2016tutorial}.

Given a tractable prior distribution $\pz$ and a likelihood $\pxz$, the marginal distribution $\px$ could be learnt by minimizing the objective
\begin{equation}
    -\log{\px}=-\log{\int_\z \pxz\pz d\z}.
\end{equation}
However, this integral is often intractable for high dimensional $\x$.
VAEs instead approximate $\px$ by minimizing the following upper bound,
\begin{equation}
\begin{split}
    &-\log{\px} + D_{KL}[\qzx \parallel \pzx] \\
    &= \mathbb{E}_{\qzx}[-\log \pxz)]+D_{KL}[\qzx)\parallel\pz],
\end{split}
\end{equation}
where $\theta$ and $\phi$ are learnable parameters and $D_{KL}$ is the always positive Kullback-Leibler (KL) divergence~\cite{kullback1951information}. Here, an approximate posterior $\qzx$ is introduced and optimized to diverge as little as possible from the true posterior $\pzx$.

The authors of DivNoising~\cite{prakash2020fully}, Hierarchical DivNoising (HDN)~\cite{prakash2022interpretable} and AutoNoise~\cite{salmon2022towards} adapt the VAE for denoising by incorporating a known explicit noise model into this objective, directing the decoder of the VAE to map the latent variable $\z$ to estimates of the signal $\s$,
\begin{equation}
\begin{split}
    &-\log{\px} + D_{KL}[\qzx \parallel \pzx] \\
    &= \mathbb{E}_{\qzx}[-\log \pxs)]+D_{KL}[\qzx)\parallel\pz],
\end{split}
\label{eq:DivNoisingLoss}
\end{equation}
where $\s=g_\theta(\z)$.

% ...................................................
\subsection{Inference in unsupervised denoising}
% ...................................................
After minimizing this new denoising objective, the signal underlying a given $\x$ is estimated by first encoding $\x$ with $\qzx$, sampling a $\z$ and mapping that sample to an estimate of the signal with $g_\theta(\z)$.
These solutions are samples from an approximation of the posterior $\psx$, which we refer to as the \emph{denoising distribution}.

Each sample from the denoising distribution is unique, allowing users to examine the uncertainty involved in their denoising problem.
However, a single consensus solution is often preferred.
The authors of~\cite{prakash2020fully, prakash2022interpretable, salmon2022towards} chose to calculate the per pixel mean of 100 or 1000 samples, deriving the minimum mean square error (MMSE) estimate of the denoising distribution, to get a consensus solution for measuring denoising performance.
Taking so many samples requires many forward passes of the denoiser and incurs a potentially prohibitive computational overhead for large datasets.

% During inference, VAE-based denoising works differently from supervised and self-supervised approaches.
% Instead of directly predicting an MMSE estimate for a noisy image, VAE-based denoisers predict a larger number of samples of possible clean images, which can then be combined by averaging to produce an approximate MMSE estimate, to achieve high quality results.
% In~\cite{prakash2022interpretable}, Prakash~\etal present an improved network architecture and show, how the system can be equipped to handle mild forms of structured noise.
% In~\cite{salmon2022towards} present an alternate approach to tackling structured noise within the VAe framework, which unfortunately cannot be applied yet in realistic settings.

% Possible solutions can be randomly sampled from this approximate posterior to provide diverse solutions to the denoising task, hence the method was named Diversity Denoising or DivNoising.
% Later, an improved version of DivNoising with an interpretable hierarchical architecture was introduced in~\cite{prakash2022interpretable}, and is known as Hierarchical DivNoising, or HDN.
% Following that, Salmon and Krull~\cite{salmon2022towards} introduced a version using autoregressive noise models for tackling structured noise, and that is known as AutoNoise.

Our method extends the high quality denoising performance and minimal training requirements of VAE-based denoisers by allowing them to directly and efficiently produce MMAE and MMSE results without repeated sampling.
% We show that the approach for structured noise presented in \cite{prakash2022interpretable}, can be directly applied in our method as well.

% ---------------------------------------------------
\section{Method}
\label{sec:method}
% ---------------------------------------------------
\figexpl
When given samples from a probability distribution, we are often interested in what a representative value of those samples is.
In the case of unsupervised denoising, we are interested in a representative image from the denoising distribution.
A common value to choose for this is the central tendency of the distribution~\cite{weisberg1992central}, a point which minimizes some measure of deviation from all of the samples.

For samples from a learned denoising distribution, $p(\hat{\s}|\x)$, over possible solutions $\hat{\s}$ for a noisy input image $\x$, this would be
% For an unsupervised denoiser that has learned a denoising distribution, $p(\hat{\s}|\x)$, over possible solutions $\hat{\s}$ for the noisy input image $\x$,
% % For a set of images $(\hat{\s}_1, \hat{\s}_2, \ldots)$ sampled from the denoising distribution of a noisy image $\x$, 
% this would be an image
\begin{equation}
    \hat{\s}^*=\arg\min_{\mathbf{y}}\mathbb{E}_{\hat{\s}|\x}[L(\mathbf{y}, \hat{\s})],
\end{equation}
where $L$ is some per-pixel loss function.
If $L$ is the $L_1$ loss,
% $L(\mathbf{y}, \hat{\s})= |\mathbf{y}-\hat{\s}|$
\begin{equation}
    L(\mathbf{y}, \hat{\s})= 1/n \sum_i^n|y_i-\hat{s}_i|,
\end{equation}
then $\hat{\s}^*$ corresponds to the pixel-wise median of the distribution, \ie, the MMAE estimate.
Here, $n$ denotes the number of pixels and $y_i$ and $\hat{s}_i$ denote $i^{\text{th}}$ pixel values.
For the $L_2$ loss,
\begin{equation}
    L(\mathbf{y}, \hat{\s})= 1/n \sum_i^n(y_i-\hat{s}_i)^2,
\end{equation}
% $L(\mathbf{y}, \hat{\s})=(\mathbf{y}-\hat{\s})^2$, 
$\hat{\s}^*$ will be the arithmetic mean, \ie, the MMSE.

The authors of~\cite{prakash2020fully, prakash2022interpretable, salmon2022towards} estimated $\hat{\s}^*$ using a large number of samples from their denoising distribution.
We propose instead training a CNN to directly predict a central tendency.

Let $h_\eta$ be our Direct Denoiser with parameters $\eta$ and $p(\hat{\s}|\x)$ be a denoising distribution.
The following objective,
\begin{equation}
    \arg\min_\eta\mathbb{E}_\x[\mathbb{E}_{\hat{\s}|\x}[L(h_\eta(\x),\hat{\s})]],
    \label{eq:optitask}
\end{equation}
where $L$ is either the $L_1$ or $L_2$ loss, would train $h_\eta$ to predict either the pixel-wise 
median or mean of $p(\hat{\s}|\x)$, respectively.
After training an unsupervised denoiser according to \cite{prakash2020fully, prakash2022interpretable, salmon2022towards}, we could train our Direct Denoiser with Eq.~\ref{eq:optitask} by sampling noisy images $\x$ from a training set and then running them through the unsupervised denoiser to obtain possible clean solutions $\hat{\s}$ from the denoising distribution.

We however find that it is possible to train both models simultaneously. 
% Let $\{\x_i\}_{i=1}^n$ be a set of $n$ noisy images and 
Let $f_{\theta, \phi}$ represent a VAE with the loss function in Equation~\ref{eq:DivNoisingLoss}, where $ \hat{\s} \sim f_{\theta, \phi}(\x)$ is a sample from the denoising distribution.

A single training step for simultaneously optimizing an unsupervised denoiser and an accompanying Direct Denoiser is as follows:
\begin{enumerate}
    \item Pass a noisy training image $\x$ to the unsupervised denoiser and sample a possible solution 
    % $f_{\theta, \phi}(\x_i)=
    $\hat{\s}$.
    \item Update the parameters $(\theta, \phi)$ towards minimizing the loss function in Equation~\ref{eq:DivNoisingLoss}.
    \item Pass the same $\x$ to the Direct Denoiser, calculating $h_\eta(\x)$.
    \item Update the parameters $\eta$ to minimize $L(h_\eta(\x), \hat{\s})$, where $L$ is the $L_1$ or $L_2$ loss function.
    \item Repeat until convergence.
\end{enumerate}

A visual representation of this training scheme can be found in Figure~\ref{fig:explainer}.

\section{Experiments}
\label{sec:experiments}
\tabpsnrs 
\figresults
Our Direct Denoiser was trained alongside HDN~\cite{prakash2022interpretable}, using six datasets of intrinsically noisy microscopy images that come with known ground truth signal.
Each dataset can be found in~\cite{prakash2022interpretable}, as can details of their size, spatial resolution and train, validation and test splits.
Note that for the \emph{Struct. Convallaria} dataset, we adapted HDN into HDN\textsubscript{3-6}, making it capable of handling structured noise.

\noindent{\textbf{Denoising Performance}}
To evaluate denoising performance, we compare the Peak Signal-to-Noise Ratio (PSNR) of our Direct Denoiser's direct solutions to the PSNR of HDN's consensus solutions.
The consensus solutions were produced by averaging samples of size 1, 10, 100 and 1000, reporting both their per-pixel median and mean.
The Direct Denoiser's solutions were reported from a network trained with an $L_1$ loss and a network trained with an $L_2$ loss.
Results are in Table~\ref{tab:PSNRs}.
Visual results from the same experiment can be seen in Figure~\ref{fig:results}.

\noindent{\textbf{Inference Times}}
We also compared inference time to denoising performance.
In Figure~\ref{fig:inferencetime}, the total time for HDN to generate 1, 10, 100 and 1000 samples for all 100 images in the \emph{Convallaria} test set was measured, then plotted against the PSNR of the mean of those samples, averaged over all 100 images.
On the same plot, the total time for our Direct Denoiser to produce single solutions for each image is plotted against their average PSNR.
Each test image consisted of 512$\times$512 pixels.

Using our GPU (an NVIDIA GeForce RTX 3090 Ti), generating a single 512$\times$512 solution from HDN's denoising distribution takes 0.076 seconds, using 2207MB of the GPU's memory.
Our Direct Denoiser takes 0.029 seconds at 1909MB to do the same.
Processing one image with either model uses the full capacity of the GPU's parallelism, so we saw no speed improvements by processing more than one image at a time.

If a consensus solution from HDN with PSNR approaching that of the the Direct Denoiser requires sampling 1000 solutions, inference with the proposed method is $2621\times$ faster.
 
\noindent{\textbf{Training Times and Memory Usage}}
Finally, the additional training time incurred by co-training HDN with the Direct Denoiser was examined.
The authors of HDN~\cite{prakash2022interpretable} train their network for 200,000 steps for all datasets, using a batch size of 64 and image patch size of 64$\times$64.
Using our GPU, training HDN alone takes 0.27 seconds per step for 15 hours total, using 13GB of GPU memory.
Training both HDN and the Direct Denoiser takes 0.34 seconds per step for 18.9 hours total, using 15GB of GPU memory.
Note that smaller virtual batches can be used as in~\cite{prakash2022interpretable} to reduce memory consumption.
For the proposed method to be a net time saving, inference would have to take 3.9 hours less.
Using our hardware and inference image resolution, time is saved when the inference test set consists of 185 images with $512\times512$ resolution.

\noindent{\textbf{Network Architecture and Training}}
The Direct Denoiser used in these experiments was a UNet~\cite{ronneberger2015u} with approximately 12 million parameters, while the unsupervised denoiser was the same Hierarchical VAE~\cite{sonderby2016ladder} used in \cite{prakash2022interpretable} with approximately 7 million parameters.
We chose to give our UNet more parameters than the Hierarchical VAE to ensure the former had the capacity to learn the full relationship between noisy images and solutions generated by the latter.
This may not have been necessary, and training a Direct Denoiser with a lower computational demand would be an interesting topic for future research.

Our UNet had a depth of four, with a residual block~\cite{he2016deep} consisting of two convolutions followed by a ReLU activation function~\cite{agarap2018deep} at each level.
Downsampling was performed by convolutions with a stride of two, and upsampling by nearest neighbor interpolation~\cite{rukundo2012nearest} followed by a single convolution with stride one.
All convolutions had a kernel size of 3.
The number of filters was 32 at the first level and that number doubled at each subsequent level.
Skip connections were merged by concatenating the skipped features with the features from the previous level and passing the two through a residual block.

Training followed the same procedure described in~\cite{prakash2022interpretable}, with the only difference being that our Direct Denoiser had its own Adamax optimizer~\cite{kingma2015adam} with an initial learning rate of 3e-4 that reduced by a factor of 0.5 when validation loss had plateaued for 10 epochs.

%.......................
\section{Discussion}
\label{sec:discussion}
%.......................
Solutions from our Direct Denoiser consistently scored a higher PSNR than consensus solutions of 1000 samples from HDN.
Table~\ref{tab:PSNRs} shows HDN's PSNRs converging towards our direct prediction result with increased sample size.
It seems that solutions from our Direct Denoiser are sometimes equivalent to averaging sample sizes orders of magnitude larger than the largest samples size we used in our experiment.
Moreover, by looking at the inference times reported in Figure~\ref{fig:inferencetime}, the time required to take such a sample size would be impractical for large datasets.

% Follow with discussion of difference in training times

%.......................
\section{Conclusions}
\label{sec:conclusions}
%.......................
We have demonstrated that an extension of the unsupervised denoising approach--the Direct Denoiser--can be used to dramatically speed up inference time, while at the same time improving performance when compared the standard inference procedure with up to 1000 sampled images.
We believe our approach will become the default way of producing central tendencies from unsupervised denoising models with the increase in speed potentially allowing an easy adaptation by the community.

While we have evaluated our method only for MSE and MAE loss functions, we believe the approach could also be used with other loss functions such as \emph{Tukey’s biweight loss}~\cite{belagiannis2015robust}, which might allow us to find regions of high probability density or even the \emph{maximum a posteriori} estimate.

Recent work in image restoration has suggested the use of more sophisticated perceptual loss functions (see \eg~\cite{mustafa2022training}).
These types of loss functions would likely only be usable in a supervised setting with clean training data and would be unlikely to work with Noise2Noise or self-supervised methods.
However, since the training targets sampled by our VAE are essentially clean images, they should be compatible with different types of complex loss functions, opening the door to using perceptual loss with noisy unpaired data.

% FUTURE WORK
% THER LOSS FUNCTIONS
% TUKEY -> MAP solution?
% PERCEPTUAL LOSS?

{\small
\bibliographystyle{ieee_fullname}
\bibliography{egbib}
}

\end{document}